\documentclass[10pt,twocolumn,letterpaper]{article}

\usepackage[pagenumbers]{wacv} 

\definecolor{wacvblue}{rgb}{0.21,0.49,0.74}
\usepackage[pagebackref,breaklinks,colorlinks,allcolors=wacvblue]{hyperref}
\usepackage{multirow}
\usepackage{booktabs}

\def\wacvPaperID{2092} % *** Enter the WACV Paper ID here
\def\confName{WACV}
\def\confYear{2026}

\title{Perception-Inspired Color Space Design for Photo White Balance Editing}

\author{
    Yang Cheng\textsuperscript{1}, 
    Ziteng Cui\textsuperscript{2}\thanks{Corresponding author}, 
    Shenghan Su\textsuperscript{1}, 
    Lin Gu\textsuperscript{3}, 
    Zenghui Zhang\textsuperscript{1} \\[0.5em] 
    \textsuperscript{1}Shanghai Jiao Tong University \quad 
    \textsuperscript{2}The University of Tokyo \quad
    \textsuperscript{3}Tohoku University \\
    {\tt\small \{yangcheng58, su2564468850, zenghui.zhang\}@sjtu.edu.cn} \\
    {\tt\small cui@mi.t.u-tokyo.ac.jp, lin@tohoku.ac.jp}
}

\begin{document}
\maketitle
\begin{abstract}
White balance (WB) is a key step in the image signal processor (ISP) pipeline that mitigates color casts caused by varying illumination and restores the scene’s true colors. Currently, sRGB-based WB editing for post-ISP WB correction is widely used~\cite{Afifi_2019_CVPR,Li2023WBFlowFW} to address color constancy failures in the ISP pipeline when the original camera RAW is unavailable. However, additive color models (e.g., sRGB) are inherently limited by fixed nonlinear transformations and entangled color channels, which often impede their generalization to complex lighting conditions.

To address these challenges, we propose a novel framework for WB correction that leverages a perception-inspired Learnable HSI (\textbf{LHSI}) color space. Built upon a cylindrical color model that naturally separates luminance from chromatic components, our framework further introduces dedicated parameters to enhance this disentanglement and learnable mapping to adaptively refine the flexibility. Moreover, a new Mamba-based network is introduced, which is tailored to the characteristics of the proposed \textbf{LHSI} color space. 

Experimental results on benchmark datasets demonstrate the superiority of our method, highlighting the potential of perception-inspired color space design in computational photography. The source code is available at  \url{https://github.com/YangCheng58/WB_Color_Space}.
\end{abstract}

\section{Introduction}
\label{sec:intro}

The Human Visual System (HVS) exhibits remarkable color constancy, maintaining stable object colors under varying illumination. In contrast, digital imaging systems lack this innate capability and depend on computational white balance (WB) algorithms to emulate it. As a fundamental component of the Image Signal Processor (ISP) pipeline, WB corrects illumination-induced color casts, restoring scene colors to their appearance under neutral lighting~\cite{Buchsbaum_1980,Van_2007,Barnard_2002}.

\begin{figure*}[htbp]
    \centering
    \includegraphics[width=1\textwidth]{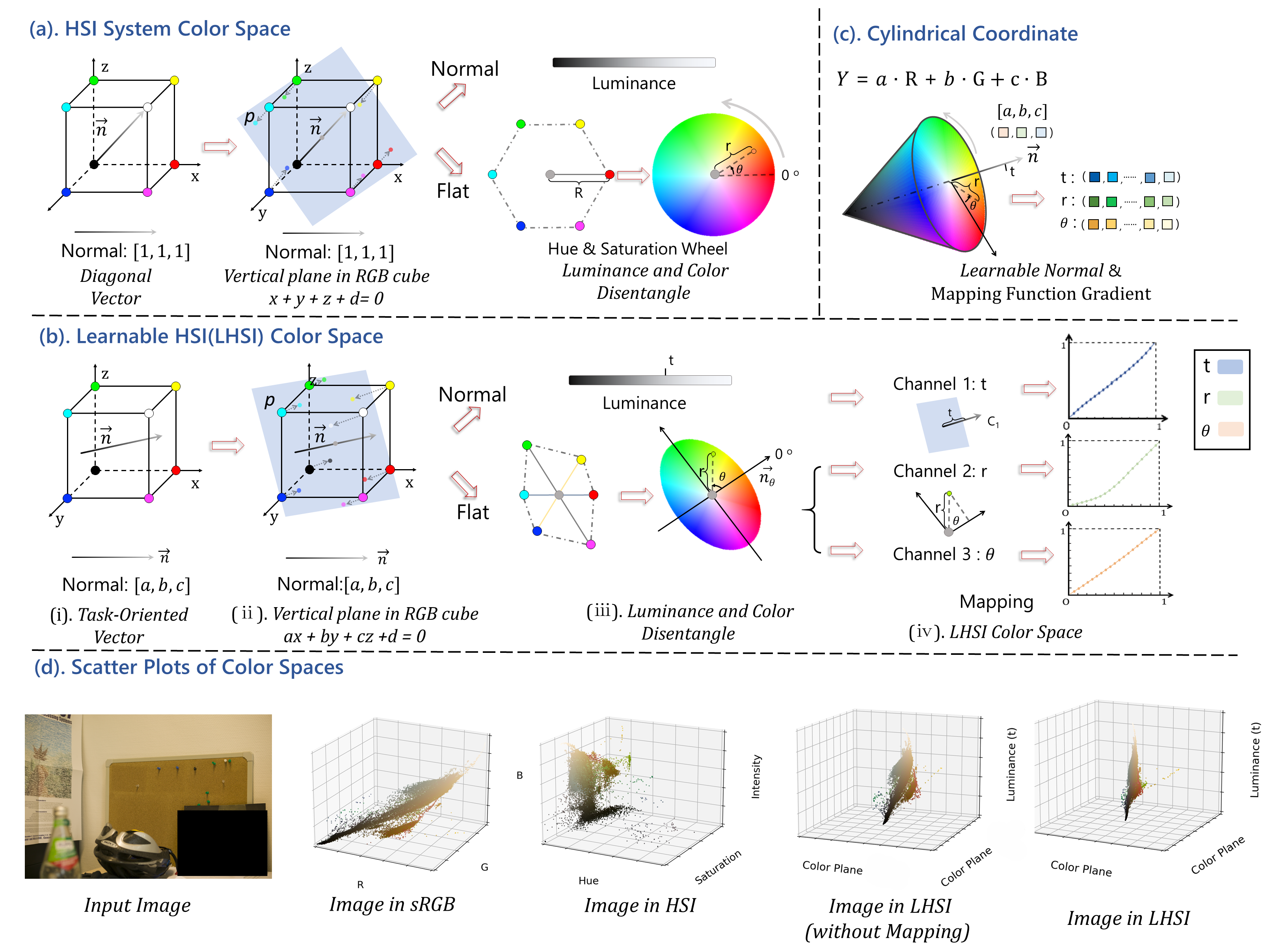}
    \vspace{-5mm}
    \caption{(a). HSI color system. (b). Our proposed Learnable HSI color space. (c) Our approach only requires learning $\overrightarrow{n}$ and Mapping Function Gradients for Hue, Saturation and Luminance axis in the model. (d).Scatter Plots of different color spaces.}
    \label{fig:colorspace}
\end{figure*}

Typically, the WB process operates on camera sensor RAW data\cite{Barron_2015_ICCV,Afifi_20201_C5,Yang_2025,HU_2017_CVPR,kim_LSMI,Xing_2022_CVPR,Chang_2025_CVPR,Wei_CVPR_2025},which linearly corresponds to the scene radiance before any nonlinear ISP operations. This linearity with respect to ambient illumination enables accurate WB estimation while avoiding interference from subsequent nonlinear ISP steps (e.g., gamma correction, 3D LUTs). However, if the in-camera WB gain is fixed and mismatched with the ambient color temperature, the resulting imbalance in RAW channels not only hinders accurate post-hoc WB correction but also leads to noticeable degradation in the final rendered sRGB image~\cite{Afifi_2019_CVPR}.

To this end, various sRGB-based WB editing methods have been proposed to post-process  sRGB images with erroneous WB settings~\cite{Afifi_2019_CVPR,Afifi_2020_CVPR,Li2023WBFlowFW,Li_AAAI_SWBNET,WACV_2023,Chiu_2025_ABCFormer,Afifi_Auto_WB}, aiming to correct illumination in the rendered sRGB space. Given the non-linear ISP pipeline, most approaches adopt end-to-end learning with deep networks. However, the sRGB color space, due to its nonlinear gamma encoding and additive channel formulation, entangles luminance and chromaticity components. This coupling makes it difficult to perform physically consistent WB adjustment, thereby limiting the robustness of color constancy estimation. Hence, here we ask: \textbf{beyond RAW and sRGB, is it possible to learn a more suitable color space for addressing the challenging problem of WB editing?}

Since RGB color spaces cannot represent uniform perceptual differences in color, more perception-oriented color spaces were developed in the 1970s to align better with human color perception. These spaces are designed to arrange colors so that changes in values are consistent with how humans perceive color differences.One such system is the HSI (Hue–Saturation–Intensity) system, which includes variations like HSB (Hue–Saturation–Brightness), HSL (Hue–Saturation–Lightness), and HSV (Hue– Saturation–Value). Another widely used system is CIELAB and its related space, CIELUV. In CIELAB, L* represents perceptual lightness, while a* and b* represent the red-green and yellow-blue color components. Similarly, in CIELUV, L* denotes lightness, with u* and v* capturing the red-green and yellow-blue color variations, respectively. 

While these spaces conceptually offer the disentanglement of luminance and chromaticity components, they rely on fixed, analytical transformations, therefore fail to account for the complex, device-specific, and often unknown non-linearities introduced by modern ISPs. Consequently, strictly applying fixed color space transformations to rendered sRGB images often yields suboptimal feature separation, limiting the efficacy of subsequent correction models.

Recent advancements have begun to challenge fixed representations by introducing learnable color spaces tailored to specific tasks. For example, Yan et al.~\cite{Yan_2025_CVPR} proposed a learnable HVI space for low-light enhancement, demonstrating that adaptive representations can capture complex illumination characteristics better than fixed formulas. Inspired by this paradigm, we propose a Learnable HSI (LHSI) color space specifically designed for sRGB white balance correction.
        
\begin{figure*}[htbp]
    \centering
    \includegraphics[width=1.0\textwidth]{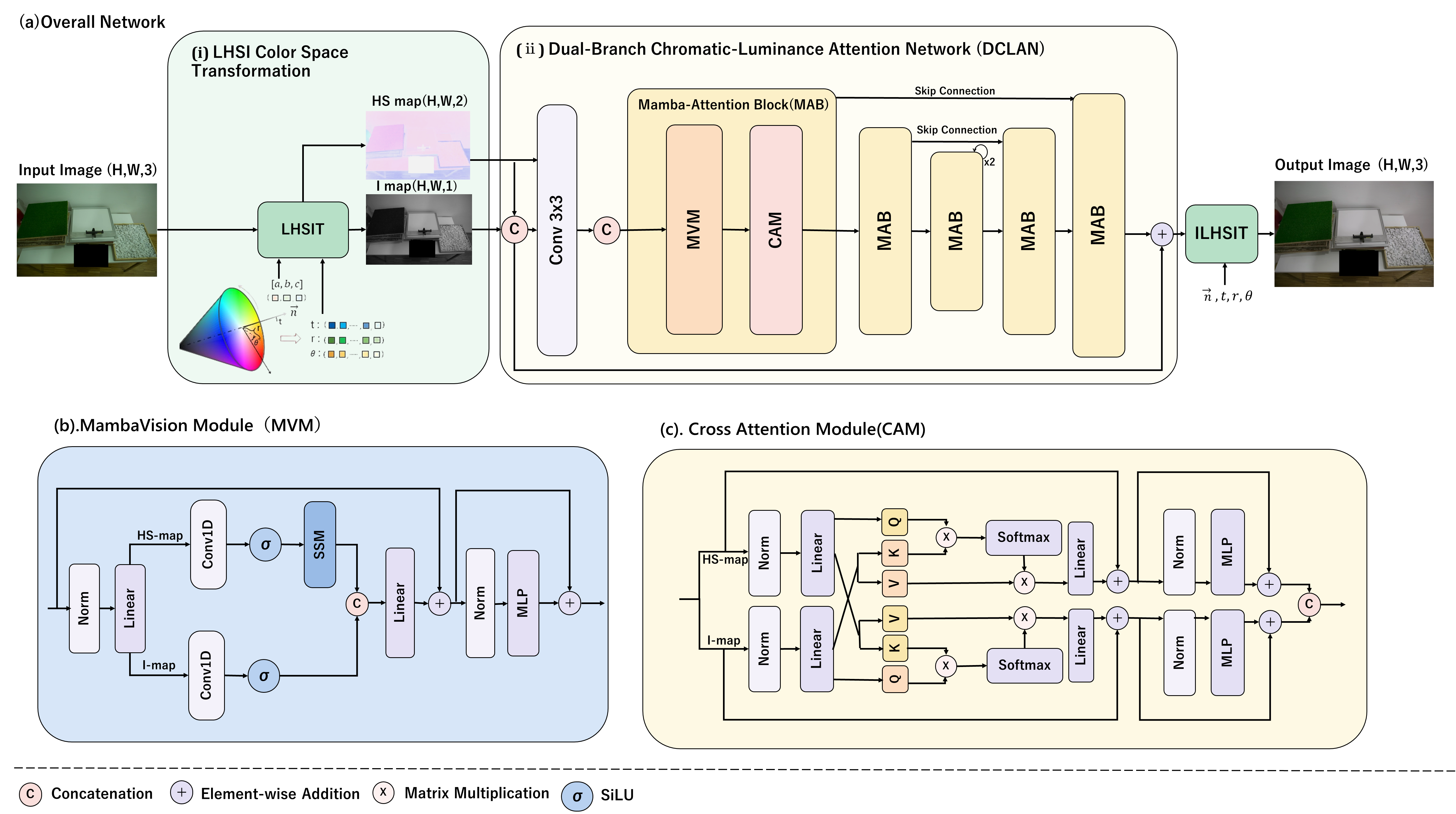}  
    \caption{(a) Overall Network. The LHSI Color Space Transformation transforms RGB images into the LHSI color space to disentangle luminance and chromaticity components. The Dual-Branch Chromatic-Luminance Attention Network (DCLAN) employs stacked Mamba-Attention Blocks (MABs) with residual connections to learn deep features before reconstructing the output via the inverse LHSI transform, which transform LHSI color space back to sRGB color space. (b) MambaVision Module (MVM). The MVM captures both long-range dependencies and local features by employing parallel branches: one utilizing a State Space Model (SSM) and the other using 1D convolution.(c) Cross Attention Module (CAM). The CAM utilizes a dual-path cross-attention mechanism to separately embed the chromatic (HS-map) and intensity (I-map) priors into the feature representations, thereby guiding the enhancement process.}  
    \label{fig:overallnetwork}
\end{figure*}

While standard HSI transformations rely on a fixed luminance axis (the diagonal of the RGB cube), the central idea of our approach is to make this axis learnable through end-to-end optimization. Our LHSI optimizes a learnable luminance axis within the network. Besides, LHSI employs nonlinear adaptive mapping functions to augment the representational flexibility. To complement this space, we have designed a specialized network that uses a MambaVision backbone~\cite{MambaVision} and a cross-attention mechanism~\cite{Yan_2025_CVPR} to process the disentangled chromatic (H, S) and intensity (I) components, aiming for improved color cast neutralization while preserving image detail.  

The main contributions of this paper are threefold:

\begin{itemize}
    \item We are the first to propose treating the color space as a learnable component in white balance correction. Our adaptive LHSI space optimizes a luminance axis and nonlinear mapping functions to guide a task-specific color space design for White Balance Correction.

    \item We developed a novel network that synergizes a MambaVision backbone with a cross-attention mechanism, specifically designed to operate within the LHSI space.

    \item Extensive experiments on multiple public WB benchmarks demonstrate that our method achieves competitive performance compared to existing sRGB-based methods in quantitative metrics, validating the potential of learnable color spaces for color constancy.
    
\end{itemize}

\section{Related Works}

\subsection{White Balance Correction}

White balance research can be broadly categorized by the data domain in which it operates: RAW or sRGB.

\textbf{RAW-based Methods}: Historically, RAW-based WB has been dominated by statistical methods that rely on priors regarding scene content. Foundational works include the Gray World assumption~\cite{Buchsbaum_1980}, which posits that the average scene reflectance is achromatic, and its variants like Gray-Edges~\cite{Van_2007}, which exploit gradient information for improved robustness. More recent statistical approaches, such as the Grayness Index (GI)~\cite{Qian_2019_CVPR}, refine these ideas by employing the dichromatic reflection model to measure the neutrality of pixels. The advent of deep learning has introduced powerful end-to-end models for illuminant estimation. 
Convolutional Color Constancy framed the problem as a learning task within a convolutional framework~\cite{Barron_2015_ICCV}, while subsequent works like C5 introduced cross-camera training strategies to enhance generalization~\cite{Afifi_20201_C5}. Recently, Chang et al.~\cite{Chang_2025_CVPR} proposed GCC, a diffusion-based approach that inpaints illumination-aligned ColorChecker charts on RAW images and extracts values from the chart’s gray patches to achieve accurate illuminant estimation. Despite their theoretical correctness, the practical utility of these RAW-based methods is often limited by the unavailability of RAW data in many user scenarios. Furthermore, applying their linear corrections directly to rendered (sRGB) images is infeasible due to the non-linear transformations inherent in the Image Signal Processor (ISP).

\textbf{sRGB-based Methods}: To address the limitations of RAW-based approaches and the ubiquity of JPEG images, a significant body of research has focused on post-stage WB correction directly in the sRGB domain. Early approaches include exemplar-based methods, which map input images to a canonical look by learning non-linear correction functions from databases of similar images~\cite{afifi2019color}. Deep learning models have further advanced this field, employing architectures such as U-Nets for end-to-end correction~\cite{Afifi_2020_CVPR} and Transformers designed to suppress color temperature-sensitive features~\cite{Li_AAAI_SWBNET}. Some approaches attempt to invert the rendering pipeline by mapping sRGB to a ``pseudo-RAW'' representation for correction before re-rendering (e.g., WBFlow~\cite{Li2023WBFlowFW}). While sRGB is convenient, its design entangles luminance and chrominance, and its non-linearity poses challenges for accurately modeling illumination changes. Recent works have thus explored diverse perspectives: Afifi et al.~\cite{afifi_2025_time-aware} leverage contextual metadata (e.g., timestamps) for mobile photography; Vasluianu et al.~\cite{vasluianu_2025_party} employ transformers to handle mixed illuminants guided by the HSV color space; and other studies~\cite{Chiu_2025_ABCFormer} enhance white balance correction by leveraging cross-modal global color knowledge from sRGB and CIELab histograms through an Interactive Channel Attention module.

\subsection{Color Spaces in Computer Vision}

Recent advancements have challenged fixed representations by introducing learnable color spaces tailored to specific tasks. For instance, Yan et al.~\cite{Yan_2025_CVPR} proposed a custom HVI (Horizontal/Vertical-Intensity) space to polarize hue-saturation maps, proving effective at suppressing color noise in dark regions. Similarly, Guan et al.~\cite{guan_2025_rethinking} developed a learnable Color Space Converter (CSC) for nighttime image deraining. By projecting RGB inputs into an adaptive YCbCr feature space, they effectively disentangled rain streaks—which are predominantly coupled with luminance in low-light environments—from complex artificial illumination, demonstrating the superiority of learned transformations over fixed linear weights. These examples highlight the paradigm shift towards utilizing task-specific color spaces, enabling better adaptability in machine learning applications ranging from low-level vision to broader fields such as robotics, industrial inspection and service robotics (e.g., housekeeping agents).

\section{Methodology}

\begin{table*}[t]
\centering
\footnotesize
\setlength{\tabcolsep}{4pt}
\caption{Quantitative comparison of different methods on three benchmark datasets: Set1-Test, Set2, and Cube+. The evaluation metrics include Mean Squared Error (MSE), Mean Absolute Error (MAE), and $\Delta$E2000. Lower values indicate better performance. The \textbf{\textit{best}}, \textbf{second-best}, and \underline{third-best} results are highlighted in the table.}
\label{tab:results}
\begin{tabular}{@{}lccccccccccccc@{}}
\toprule
\multirow{2}{*}{Method} & \multicolumn{4}{c}{MSE $\downarrow$} & \multicolumn{4}{c}{MAE (°) $\downarrow$} & \multicolumn{4}{c}{$\Delta$E2000 $\downarrow$} & \multirow{2}{*}{Size (MB) $\downarrow$} \\
\cmidrule(lr){2-5} \cmidrule(lr){6-9} \cmidrule(lr){10-13}
 & Mean & Q1 & Q2 & Q3 & Mean & Q1 & Q2 & Q3 & Mean & Q1 & Q2 & Q3 & \\
\midrule
\multicolumn{14}{c}{Rendered WB Dataset: Set1-Test (21,046 images) [3]} \\
\midrule
KNN ~\cite{Afifi_2019_CVPR} & \underline{77.49} & 13.74 & \underline{39.62} & \underline{94.01} & \underline{3.06} & \underline{1.74} & \underline{2.54} & \underline{3.76} & \underline{3.58} & \underline{2.07} & \underline{3.09} & \underline{4.55} & 21.8 \\
Deep-WB ~\cite{Afifi_2020_CVPR} & 82.55 & \underline{13.19} & 42.77 & 102.09 & 3.12 & 1.88 & 2.70 & 3.84 & 3.77 & 2.16 & 3.30 & 4.86 & \underline{16.7} \\
Mixed-WB ~\cite{Afifi_Auto_WB} & 142.25 & 26.81 & 67.17 & 164.66 & 4.07 & 2.64 & 3.68 & 5.16 & 4.55 & 3.00 & 4.45 & 5.63 & \textbf{\textit{5.1}} \\
WBFlow ~\cite{Li2023WBFlowFW} & \textbf{78.89} & \textbf{12.99} & \textbf{35.09} & \textbf{79.35} & \textbf{2.67} & \textbf{1.73} & \textbf{2.39} & \textbf{3.24} & \textbf{3.13} & \textbf{1.92} & \textbf{2.79} & \textbf{3.94} & 30.2 \\
SWBNet ~\cite{Li_AAAI_SWBNET} & 111.62 & 20.61 & 60.68 & 137.91 & 4.11° & 2.56° & 3.75° & 5.22° & 4.54 & 2.73 & 4.16 & 5.86 & 258.8 \\
Ours & \textbf{\textit{54.79}} & \textbf{\textit{10.52}} & \textbf{\textit{26.60}} & \textbf{\textit{59.04}} & \textbf{\textit{2.53}} & \textbf{\textit{1.59}} & \textbf{\textit{2.26}} & \textbf{\textit{3.13}} & \textbf{\textit{2.79}} & \textbf{\textit{1.74}} & \textbf{\textit{2.43}} & \textbf{\textit{3.42}} & \textbf{6.40} \\
\midrule
\multicolumn{14}{c}{Rendered WB Dataset: Set2 (2,881 images) [3]} \\
\midrule
KNN ~\cite{Afifi_2019_CVPR} & 171.09 & 37.04 & 87.04 & 190.88 & 4.48 & 2.26 & 3.64 & 5.95 & 5.60 & 3.43 & 4.90 & 7.06 & 21.8 \\
Deep-WB ~\cite{Afifi_2020_CVPR} & \underline{124.07} & \textbf{30.13} & \underline{76.32} & \underline{154.44} & \underline{3.75} & \underline{2.02} & \underline{3.08} & \underline{4.72} & \underline{4.90} & \textbf{3.13} & \underline{4.35} & \underline{6.08} & \underline{16.7} \\
Mixed-WB ~\cite{Afifi_Auto_WB} & 188.76 & 48.64 & 112.32 & 219.91 & 4.92 & 2.69 & 4.10 & 6.37 & 6.05 & 3.45 & 4.92 & 7.20 & \textbf{\textit{5.1}} \\
WBFlow ~\cite{Li2023WBFlowFW} & \textbf{117.60} & \underline{31.25} & \textbf{\textit{61.68}} & \textbf{143.90} & \textbf{\textit{3.51}} & \textbf{\textit{1.93}} & \textbf{2.92} & \textbf{4.47°} & \textbf{4.64} & \textbf{3.13} & \textbf{\textit{4.07}} & \textbf{5.56} & 30.2 \\
SWBNet ~\cite{Li_AAAI_SWBNET} & 219.02 & 55.45 & 113.98 & 236.25 & 5.46 & 3.45 & 4.78 & 6.63 & 6.51 & 4.39 & 5.84 & 8.08 & 258.8 \\
Ours & \textbf{\textit{115.53}} & \textbf{\textit{29.97}} & \textbf{64.63} & \textbf{\textit{143.34}} & \textbf{3.56} & \textbf{1.95} & \textbf{\textit{2.81}} & \textbf{\textit{4.44}} & \textbf{\textit{4.55}} & \textbf{\textit{3.05}} & \textbf{4.14} & \textbf{\textit{5.55}} & \textbf{6.40} \\
\midrule
\multicolumn{14}{c}{Rendered Cube+ Dataset (10,242 images) [3, 6]} \\
\midrule
KNN ~\cite{Afifi_2019_CVPR} & 194.98 & 27.43 & 57.08 & 118.21 & 4.12 & 1.96 & 3.17 & 5.04 & 5.68 & 3.22 & 4.61 & 6.70 & 21.8 \\
Deep-WB ~\cite{Afifi_2020_CVPR} & 80.46 & \textbf{15.43} & \underline{33.88} & \textbf{74.42} & \underline{3.45} & \underline{1.87} & \underline{2.82} & \underline{4.26} & 4.59 & \underline{2.68} & 3.81 & 5.53 & \underline{16.7} \\
Mixed-WB ~\cite{Afifi_Auto_WB} & 161.80 & \underline{16.96} & \textbf{\textit{19.33}} & 90.81 & 4.05 & \textbf{1.40} & \textbf{2.12} & 4.88 & 4.89 & \textbf{\textit{2.16}} & \textbf{\textit{3.10}} & 6.78 & \textbf{\textit{5.1}} \\
WBFlow ~\cite{Li2023WBFlowFW} & \textbf{75.39} & \textbf{\textit{14.22}} & \textbf{30.90} & \textbf{\textit{72.91}} & \textbf{3.34} & \underline{1.87} & \underline{2.82} & \textbf{\textit{4.11}} & \textbf{\textit{4.28}} & \underline{2.68} & \underline{3.77} & \textbf{5.21} & 30.2 \\
SWBNet ~\cite{Li_AAAI_SWBNET} & \textbf{\textit{74.35}} & 20.46 & \underline{40.04} & \underline{86.95} & \textbf{\textit{3.15}} & \textbf{\textit{1.33}} & \textbf{\textit{2.09}} & \textbf{4.12} & \textbf{\textit{4.28}} & \textbf{2.40} & \textbf{3.16} & \textbf{\textit{5.09}} & 258.8 \\
Ours & \underline{78.19} & 23.73 & 44.67 & 87.27 & 3.70 & 2.14 & 3.11 & 4.51 & \underline{4.41} & 2.81 & 3.88 & \underline{5.41} & \textbf{6.40} \\
\bottomrule
\end{tabular}
\end{table*}

\subsection{Learnable HSI Color Space}

For the color space translation part, we extend the popular cylindrical-coordinate design used for HSI system. As shown in \cref{fig:colorspace} (a), HSI system at first fixed the luminance axis to be a diagonal vector from black (0, 0, 0) to white (1, 1, 1) in the RGB cube. The projection of any RGB value is then defined as the lightness or brightness representing the physical illumination component. The projection plane of the diagonal line is taken as the luminance-independent chromatic plane $\mathcal{P}$, where in chromatic plane $\mathcal{P}$, the angle around the luminance axis corresponds to \textit{Hue} and the distance from the axis corresponds to \textit{Saturation}. 

The luminance axis in a 3D RGB cube is defined differently across various color spaces~\cite{Nguyen_2017_CVPR,Color_to_Grayscale}, such as $Y = \frac{1}{3} \cdot (R + G + B)$ in the HSI color space and $Y = 0.299 \cdot R + 0.587 \cdot G + 0.114 \cdot B$
in the YUV color space. In our color system, we aim to establish an objective-driven luminance axis $Y  = \boldsymbol{a} \cdot R + \boldsymbol{b} \cdot G + \boldsymbol{c} \cdot B$ within the 3D RGB cube, capable of adapting to computer vision tasks under challenging lighting conditions.

A pixel of position $(x,y)$ in image $I(x,y)$  can be projected to new luminance axis with direction $\overrightarrow{n}=(a,b,c),\| \overrightarrow{n} \|_2 = 1$, with values $t(x,y)$ calculated by $a\cdot R(x,y)+b\cdot B(x,y)+c\cdot G(x,y)$ and then normalized to between 0-1.

The chromatic component of the pixel, independent of luminance, is obtained by projecting $I(R,G,B)$ onto the plane orthogonal to the luminance axis $\vec{n}$. The projection coordinates on this plane are represented in polar form as $\text{Proj}_{\perp \vec{n}} I(x, y) = (r(x, y), \theta(x, y))$, where $r(x, y)$ is the radial distance from the luminance axis, denoting the saturation, and $\theta(x, y)$ is the angular position, representing the hue, the hue and saturation components are then normalized to between 0-1. 

To augment the representational capacity of the proposed task-specific color space, we incorporate non-linear mapping strategies across all three component channels. To go into detail, we define the mapping as a monotonic piecewise linear function. Given an input variable $v \in [0, 1]$ (representing $t$,$r$ or $\theta$), we discretize the domain into $M$ uniform intervals of length $\delta = 1/M$. To model the local stretching or compressing of the colour space, we assign a learnable parameter $u_i$ to the $i$-th interval. The slope $\alpha_i$ for each interval is constrained to be positive to strictly guarantee monotonicity and invertibility:
\begin{equation}
    \alpha_i = \alpha_{min} + (\alpha_{max} - \alpha_{min}) \cdot \sigma(u_i), \quad i \in \{0, \dots, M-1\},
\end{equation}

where $\sigma(\cdot)$ is the sigmoid function, and $[\alpha_{min}, \alpha_{max}]$ constraint the learnable range of the slopes. For an input value $v$ falling into the $k$-th interval (where $k = \lfloor v / \delta \rfloor$), the mapped output is calculated by accumulating the contributions of previous intervals plus the partial contribution within the current interval. To ensure the output remains within $[0, 1]$, we normalize the result by the total integral of the function:
\begin{equation}
\begin{aligned}
    \mathcal{M}_{v}(v) &= \frac{1}{Z} \left( \sum_{i=0}^{k-1} \alpha_i \delta\right) + \alpha_k (v - k\delta) ,v\in(t,r,\theta)
\end{aligned}
\end{equation}

where $Z$ acts as the normalization factor, as $Z = \sum_{i=0}^{M-1} \alpha_i \delta$.

This formulation ensures that $\mathcal{M}(\cdot)$ is a bijection. Consequently, the original values can be losslessly recovered via its inverse mapping, which is essential for reconstructing the RGB image from the LHSI representation. After the nonlinear mapping we now have $t'=M_{t}(t)$, $t=r'=M_{r}(r)$ and $\theta'=M_{\theta}(\theta)$. Inspired by the HVI color space~\cite{Yan_2025_CVPR}, the resulting representation, composed of $(t',r'\cdot sin(\theta'),r'\cdot cos(\theta'))$, forms the basis of the proposed LHSI color space.

\subsection{Dual-Branch Chromatic-Luminance Attention Network (DCLAN)}
After The LHSI color space transformation, the input sRGB images are divided into intensity(I)-maps and chromatic (HS)-maps. Inspired by ~\cite{Yan_2025_CVPR}, we concat HS-maps and I-maps for the I branch,while HS-map for the HS-branch, and send them to the proposed U-net like Dual-Branch Chromatic-Luminance Attention Network (DCLAN) ,which adopts a hierarchical Encoder-Decoder architecture designed to effectively enhance the intensity and chromaticity components of images under incorrect white balance settings.

\subsubsection{Mamba-Attention Block (MAB)}
 As illustrated in \cref{fig:overallnetwork}(a), to effectively capture deep feature representations, we design the Mamba-Attention Block (MAB) as the fundamental building block of our DCLAN. It comprises two distinct components: the MambaVision Module (MVM), which exploits the State Space Model to model long-range dependencies efficiently, and the Cross Attention Module (CAM), which incorporates the disentangled LHSI priors for precise feature refinement. The details of these two components are elaborated below.

\begin{figure*}[htbp]
    \centering
    \includegraphics[width=1.0\textwidth]{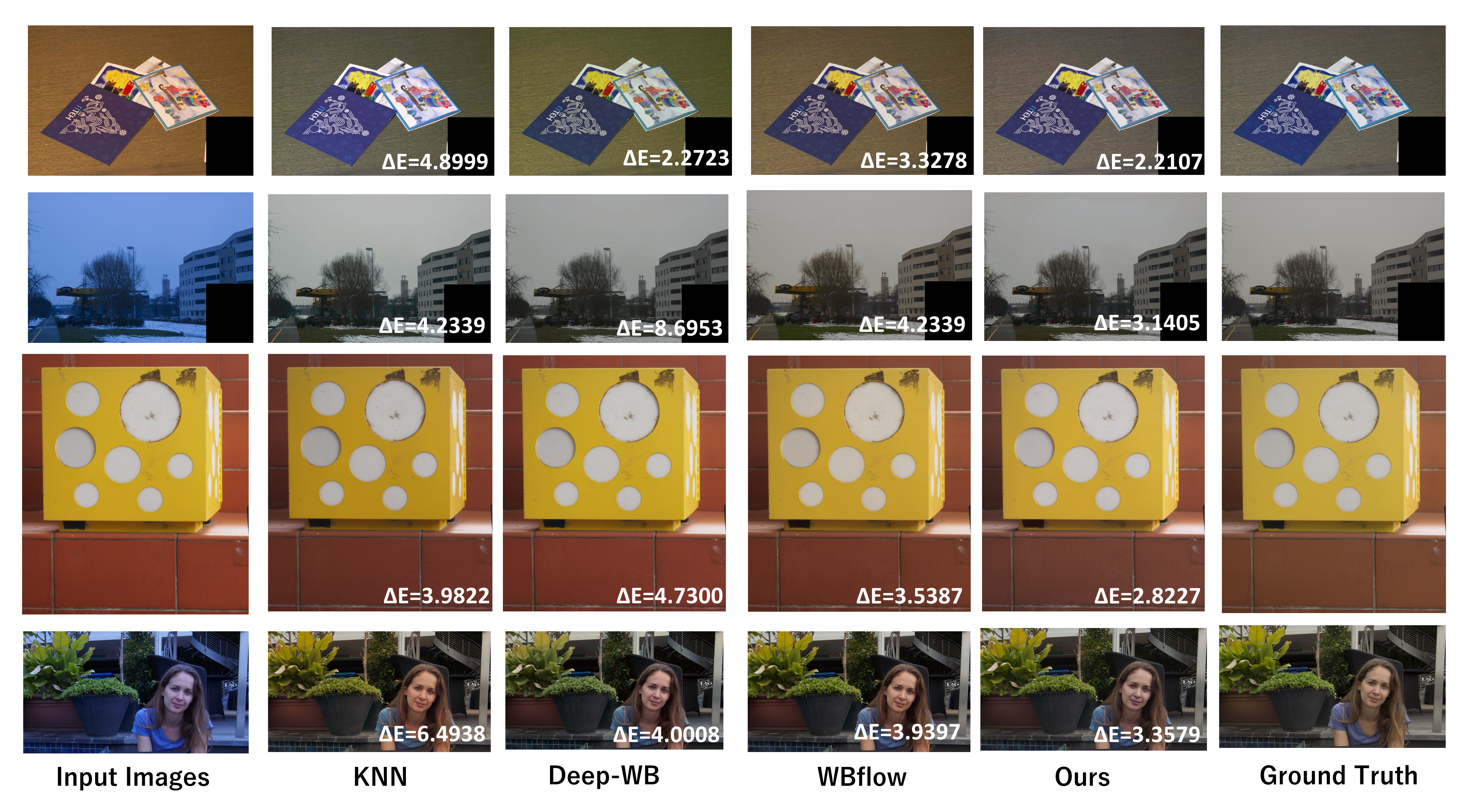}  
    \caption{Visual comparison of white balance correction results. From left to right: input image, results from KNN, Deep-WB, WBFlow, our method, and the ground truth. $\Delta$E2000 values indicate the perceptual difference from the ground truth.}
    \label{fig:results}
\end{figure*}

\subsubsection{MambaVision Module (MVM)}
\label{sec:mvm}

To effectively correct the color cast, the network must capture global illumination statistics while preserving local structural details. We incorporate the MambaVision Module (MVM), which leverages Structured State Space Models (SSMs) to model long-range dependencies with linear complexity ($\mathcal{O}(N)$). As illustrated in ~\cref{fig:overallnetwork}(b), our MVM is specifically designed to process visual signals by integrating local convolution with global selective scanning in a dual-branch architecture.

\textbf{Selective State Space Model}:
The core of the MVM is the continuous-time system that maps a 1D input $x(t) \in \mathbb{R}$ to an output $y(t) \in \mathbb{R}$ through a latent state $h(t) \in \mathbb{R}^N$:
\begin{equation}
    h'(t) = \mathbf{A}h(t) + \mathbf{B}x(t), \quad y(t) = \mathbf{C}h(t).
\end{equation}
To implement this in discrete-time deep learning, the parameters $(\mathbf{A}, \mathbf{B})$ are discretized using the Zero-Order Hold (ZOH) principle with a learnable timescale parameter $\Delta$. The discretized parameters $\overline{\mathbf{A}} = \exp(\Delta \mathbf{A})$ and $\overline{\mathbf{B}} = (\Delta \mathbf{A})^{-1}(\exp(\Delta \mathbf{A}) - \mathbf{I}) \cdot (\Delta \mathbf{B})$ allow the system to be computed recursively:
\begin{equation}
    h_t = \overline{\mathbf{A}} h_{t-1} + \overline{\mathbf{B}} x_t, \quad y_t = \mathbf{C} h_t.
\end{equation}
Crucially, the parameters $\Delta, \mathbf{B}, \mathbf{C}$ are derived from the input $x_t$ (i.e., they are content-aware), enabling the Selective Scan mechanism that dynamically filters information based on the input context.

\textbf{Proposed Network}:
Based on our specific implementation, the MVM processes the input features $\mathbf{X}_{in}$ through a parallel workflow. First, the input is passed through layer normalization modules and projected to an expanded dimension and split along the channel dimension into two branches: the intensity branch ($F_{\mathbf{I}}$) and the Hue-Saturation branch($F_{\mathbf{HS}}$).

To inject inductive biases for local pixel dependencies, both branches are processed by a depth-wise 1D convolution followed by a SiLU activation:

\begin{equation}
\begin{aligned}
    F_{\mathbf{I}}' &= \text{SiLU}(\text{DWConv1d}(F_{\mathbf{I}})), \\
    F_{\mathbf{HS}}' &= \text{SiLU}(\text{DWConv1d}(F_{\mathbf{HS}})).
\end{aligned}
\end{equation}

Subsequently, only the HS Branch is processed by the Selective Scan Mechanism (SSM) to capture long-range interactions, while the I Branch retains fine-grained spatial details unmodified by the state space. Finally, the features from both branches are concatenated and fused via a linear projection layer:
\begin{equation}
    F_{\mathbf{HS}_{ssm}} = \text{SSM}(F_{\mathbf{HS}}'), 
\end{equation}
\begin{equation}
    \mathbf{X}_{ssm} = \text{Linear}\left( \text{Concat}[F_{\mathbf{I}}', F_{\mathbf{HS}_{SSM}}] \right). 
\end{equation}

Subsequently, the features are passed through a MLP layer for further extraction. Simultaneously,  residual connections are incorporated.  

\begin{equation}
\mathbf{X}_{ssm}'=(\mathbf{X}_{in}+\mathbf{X}_{ssm}),
\end{equation}
\begin{equation}
\mathbf{X}_{out}=\mathbf{X}_{ssm}'+MLP[Norm(\mathbf{X}_{ssm}')].
\end{equation}

\subsection{Cross-Attention Module (CAM)}
As illustrated in \cref{fig:overallnetwork}(c), the input features are first partitioned along the channel dimension into two distinct branches: chromatic features $\mathbf{HS}_{in}$ and intensity features $\mathbf{I}_{in}$. To effectively fuse these disentangled representations, the CAM employs a cross-attention mechanism where one branch serves as the query to retrieve complementary context from the other. Specifically, for the intensity branch update, the query $\mathbf{Q}_{I}$ is projected from the intensity features $\mathbf{X}_{I}$, while the keys $\mathbf{K}_{HS}$ and values $\mathbf{V}_{HS}$ are derived from the chromatic features $\mathbf{X}_{HS}$:
\begin{equation}
    \mathbf{Q}_{I} = \mathbf{I}_{in}\mathbf{W}^Q, \quad \mathbf{K}_{HS} = \mathbf{HS}_{in}\mathbf{W}^K, \quad \mathbf{V}_{HS} = \mathbf{HS}_{in}\mathbf{W}^V,
\end{equation}

where $\mathbf{W}^Q, \mathbf{W}^K, \mathbf{W}^V$ denote the learnable projection matrices.

The attention map is computed via scaled dot-product:

\begin{equation}
    I_{Attn} = Linear\left(\text{Softmax}\left(\frac{\mathbf{Q}_{I}\mathbf{K}_{HS}^T}{\sqrt{d_k}}\right)\mathbf{V}_{HS}\right).
\end{equation}

Subsequently, the features are passed through a MLP layer for further extraction. Simultaneously,  residual connections are incorporated.  

\begin{equation}
     \mathbf{I}_{Attn}'=(\mathbf{I}_{Attn}+\mathbf{I}_{in}),
\end{equation}
\begin{equation}
\mathbf{I}_{out}=\mathbf{I}_{Attn}'+MLP[Norm(\mathbf{I}_{Attn}')]
\end{equation}

This operation aligns the structural details of the intensity map with the color consistency of the chromatic map. The HS branch shares an identical architecture.
\section{Experiments and Results}

\subsection{Datasets}

For our experiments, we reference ~\cite{Afifi_2020_CVPR} and build our training set on Rendered WB
dataset ~\cite{Afifi_2019_CVPR} by randomly sampling 12,000 rendered sRGB images with diverse white balance settings from two of the three non-overlapping folds of the Set1 partition.

To evaluate performance, we utilize the third fold of Set1 of Rendered WB
dataset, designated as Set1-Test (21,046 images)~\cite{Afifi_2020_CVPR}, as our primary in-distribution test set. Furthermore, to rigorously assess the generalization capabilities of our method on unseen scenes and camera models, we benchmark its performance on two additional out-of-distribution datasets: Set2 of the Rendered WB dataset (2,881 images)~\cite{Afifi_2020_CVPR}  and the Rendered Cube+ dataset (10,242 images) ~\cite{Afifi_2020_CVPR}.

\subsection{Experiment settings}
Our model was implemented using the PyTorch framework, and all experiments were conducted on a workstation equipped with an NVIDIA Quadro RTX 6000 GPU. During training, input images were randomly cropped to a resolution of 256x256 pixels. To optimize the network, we utilized the Mean Absolute Error (MAE) loss$L$. The total loss function is defined as a weighted sum of the RGB loss and the LHSI loss, specifically $0.9 L_{RGB} + 0.1 L_{LHSI}$.The network was trained using the Adam optimizer ($\beta_1=0.9, \beta_2=0.99$) with an initial learning rate of $1 \times 10^{-4}$ and weight decay set to 0. We utilized a MultiStepLR scheduler to decay the learning rate by reducing the learning rate by a factor of 0.5 at epochs 40, 70, 90 and 110.

The above datasets contain images of varying sizes. To ensure consistency, all input images to the network are resized to 256×256. Following Afifi et al.'s method ~\cite{afifi2019color}, we apply the learned mapping functions to generate corrected images at their original resolutions. This involves resizing the corrected small images back to the target image dimensions, enabling direct comparison with ground truth images. This approach ensures a fair and accurate evaluation of our method's performance on diverse image resolutions.

We follow the same evaluation metrics
used by ~\cite{Afifi_2020_CVPR}. Specifically, we used
the following metrics to evaluate our results: mean square
error (MSE), mean angular error (MAE), and $\Delta$ E 2000~\cite{Sharma2005TheCC}. For each evaluation metric, we report the mean, lower
quartile (Q1), median (Q2), and the upper quartile (Q3) of the error.

\subsection{Results and Discussion}

After 120 epochs of training, the direction $\overrightarrow{n}=(a,b,c),\| \overrightarrow{n} \|_2 = 1$ is $\overrightarrow{n}=(0.48, 0.76,0.45)$.The mapping functions are shown in \cref{fig:colorspace}(b).  

As shown in \cref{tab:results}, our method achieves competitive performance on the Rendered WB Set1 dataset with a Mean MSE of 54.79. On datasets with differing distributions (Set2 and Cube+), our model maintains results comparable to leading approaches, though it does not consistently rank highest. We attribute this to the specialized design of our LHSI color space and DCLAN architecture: while leveraging domain-specific priors enhances in-distribution accuracy, it may inherently constrain broader generalization compared to other models. Furthermore, our model size is extremely compact (\textbf{6.40 MB}, $\sim$21\% of WBFlow).

The  results in the \cref{fig:results} illustrate the effectiveness of our method in suppressing color deviations compared to other approaches. The results indicates closer alignment with the ground truth across diverse scenes. These results, combined with the quantitative metrics, demonstrate the robustness and perceptual accuracy of our approach.

\subsection{Experiments on different color spaces}
\label{sec:ablation}  

To validate the effectiveness of the proposed LHSI space, we conducted a comprehensive ablation study comparing it against standard color spaces (CIELAB, HSV, and HVI). To ensure fair comparison, input channels for all spaces were normalized to comparable ranges (e.g., $L^*$ in CIELAB to $[0,1]$, and $a^*, b^*$ to $[-1, 1]$).

\paragraph{Performance with U-Net}
We first established a baseline using a standard U-Net architecture to evaluate the intrinsic representation capability of each color space in isolation, independent of our DCLAN architecture. The models were trained for 100 epochs using the Adam optimizer (initial $lr=1\times10^{-4}$, weight decay $10^{-5}$, $\beta_1=0.9, \beta_2=0.999$)with a Mean Absolute Error (MAE) loss calculated in the sRGB domain. The learning rate was decayed by a factor of 0.5 every 25 epochs.

After training, the direction of LHSI color space $\overrightarrow{n}=(a,b,c),\| \overrightarrow{n} \|_2 = 1$ is $\overrightarrow{n}=(0.55,0.60,0.59)$.
The quantitative results, summarized in \cref{tab:colorspace_unet}, indicate that the LHSI space consistently outperforms alternatives across all three datasets. This suggests that LHSI provides a more  effective representation for the network, facilitating better convergence even under a simple regression objective.

\begin{table}[t]
\centering
\footnotesize
\caption{Quantitative comparison of different color spaces using a standard U-Net baseline. \textbf{Bold} indicates the best performance.}
\label{tab:colorspace_unet}
\setlength{\tabcolsep}{3.5pt}
\begin{tabular}{@{}lcccc@{}}
\toprule
\multirow{2}{*}{Method} & \multirow{2}{*}{Dataset} & MSE & MAE ($^{\circ}$) & $\Delta$E2000 \\
 & & Mean $\downarrow$ & Mean $\downarrow$ & Mean $\downarrow$ \\
\midrule
CIELAB & \multirow{4}{*}{Set1-test} & 118.80 & 4.39 & 4.57 \\
HSV & & 113.91 & 4.30 & 4.46 \\
HVI & & 101.79 & 3.87 & 4.17 \\
Ours (LHSI) & & \textbf{71.54} & \textbf{3.09} & \textbf{3.42} \\
\midrule
CIELAB & \multirow{4}{*}{Set2} & 216.33 & 5.32 & 6.51 \\
HSV & & 206.31 & 5.13 & 6.27 \\
HVI & & 204.68 & 4.95 & 6.14 \\
Ours (LHSI) & & \textbf{135.11} & \textbf{3.97} & \textbf{5.14} \\
\midrule
CIELAB & \multirow{4}{*}{Cube+} & 130.97 & 4.63 & 5.33 \\
HSV & & 120.27 & 4.30 & 4.97 \\
HVI & & 120.81 & 4.32 & 5.03 \\
Ours (LHSI) & & \textbf{87.57} & \textbf{3.74} & \textbf{4.39} \\
\bottomrule
\end{tabular}
\end{table}

\paragraph{Performance within the DCLAN Framework.}
We further evaluated these color spaces integrated into our full DCLAN framework. The experimental setup remained identical to the baseline to isolate the contribution of the color space.

As observed in \cref{tab:dclan_comparison}, our proposed LHSI space enables the DCLAN framework to achieve superior perceptual quality compared to CIELAB, HSV, and HVI. Unlike fixed coordinate systems, LHSI adaptively aligns the illumination axis according to the white balance correction task. This adaptability allows the network to fit the color distribution of the ground truth more accurately, resulting in lower error rates across most metrics.

\section{Conclusion}

\begin{table}[t]
\centering
\footnotesize
\caption{Quantitative comparison of different color spaces using the full DCLAN framework. \textbf{Bold} indicates the best performance.}
\label{tab:dclan_comparison}
\setlength{\tabcolsep}{3.5pt}
\begin{tabular}{@{}lcccc@{}}
\toprule
\multirow{2}{*}{Method} & \multirow{2}{*}{Dataset} & MSE & MAE ($^{\circ}$) & $\Delta E_{2000}$ \\
 & & Mean $\downarrow$ & Mean $\downarrow$ & Mean $\downarrow$ \\
\midrule
CIELAB & \multirow{4}{*}{Set 1-test} & 64.79 & 2.80 & 3.11 \\
HSV & & 61.06 & 2.87 & 3.12 \\
HVI & & 62.08 & 2.86 & 3.09 \\
Ours (LHSI) & & \textbf{54.79} & \textbf{2.53} & \textbf{2.79} \\
\midrule
CIELAB & \multirow{4}{*}{Set 2} & 121.88 & 3.70 & 4.72 \\
HSV & & 121.48 & 3.81 & 4.88 \\
HVI & & 121.43 & 3.76 & 4.80 \\
Ours (LHSI) & & \textbf{115.53} & \textbf{3.56} & \textbf{4.55} \\
\midrule
CIELAB & \multirow{4}{*}{Cube+} & 90.43 & 3.75 & 4.60 \\
HSV & & 85.97 & \textbf{3.66} & 4.45 \\
HVI & & 82.70 & 3.75 & 4.47 \\
Ours (LHSI) & & \textbf{78.19} & 3.70 & \textbf{4.41} \\
\bottomrule
\end{tabular}
\end{table}

In this paper, we have presented a novel framework for post-ISP white balance correction. Central to our method is the Learnable HSI (LHSI) color space, which departs from fixed color space transforms by employing a data-driven strategy to explicitly disentangle luminance from chromatic components. To fully exploit this representation, we introduced a specialized architecture tailored to process these decoupled features, enabling precise chromatic alignment while preserving structural details.

Extensive experiments demonstrate that our approach achieves competitive results compared to other state-of-the-art methods and outperforms standard color spaces (e.g., CIELAB, HVI). While the strong, task-specific priors of our learned space provide superior performance on target distributions, they also present a trade-off regarding zero-shot generalization across unseen camera sensors. Future work will focus on enhancing the cross-sensor adaptability of the LHSI representation. Nevertheless, this work underscores the critical role of adaptive color space design in computational photography and establishes a promising direction for learning-based ISP enhancement.
{
    \small
    \bibliographystyle{ieeenat_fullname}
    \bibliography{main}
}

\end{document}